\documentclass{vgtc}                          % final (conference style)
\ifpdf%                                % if we use pdflatex
  \pdfoutput=1\relax                   % create PDFs from pdfLaTeX
  \usepackage{graphicx}                % allow us to embed graphics files
  \DeclareGraphicsExtensions{.pdf,.png,.jpg,.jpeg} % for pdflatex we expect .pdf, .png, or .jpg files
\else%                                 % else we use pure latex
  \usepackage{graphicx}                % allow us to embed graphics files
  \DeclareGraphicsExtensions{.eps}     % for pure latex we expect eps files
\fi%

\graphicspath{{figures/}{pictures/}{images/}{./}} % where to search for the images

\usepackage{microtype}                 % use micro-typography (slightly more compact, better to read)
\PassOptionsToPackage{warn}{textcomp}  % to address font issues with \textrightarrow
\usepackage{textcomp}                  % use better special symbols
\usepackage{mathptmx}                  % use matching math font
\usepackage{times}                     % we use Times as the main font
\usepackage{cite}                      % needed to automatically sort the references
\usepackage{tabu}                      % only used for the table example
\usepackage{booktabs}                  % only used for the table example
\onlineid{1065}

\vgtccategory{Application}

\vgtcinsertpkg

\title{HAiVA: Hybrid AI-assisted Visual Analysis Framework to Study the Effects of Cloud Properties on Climate Patterns}

\author{Subhashis Hazarika$^1$\thanks{e-mail: shazarika@parc.com} %
\and Haruki Hirasawa$^2$ %
\and Sookyung Kim$^1$ %
\and Kalai Ramea$^1$ %
\and Salva R. Cachay$^4$ %
\and Peetak Mitra$^3$ %
\and Dipti Hingmire$^2$ %
\and Hansi Singh$^2$ %
\and Phil J. Rasch$^5$
}
\affiliation{\scriptsize $^1$Palo Alto Research Center, $^2$University of Victoria, $^3$Excarta, $^4$University of California, San Dieago , $^5$University of Washington }

\teaser{
  \centering
  \includegraphics[width=1.0\linewidth]{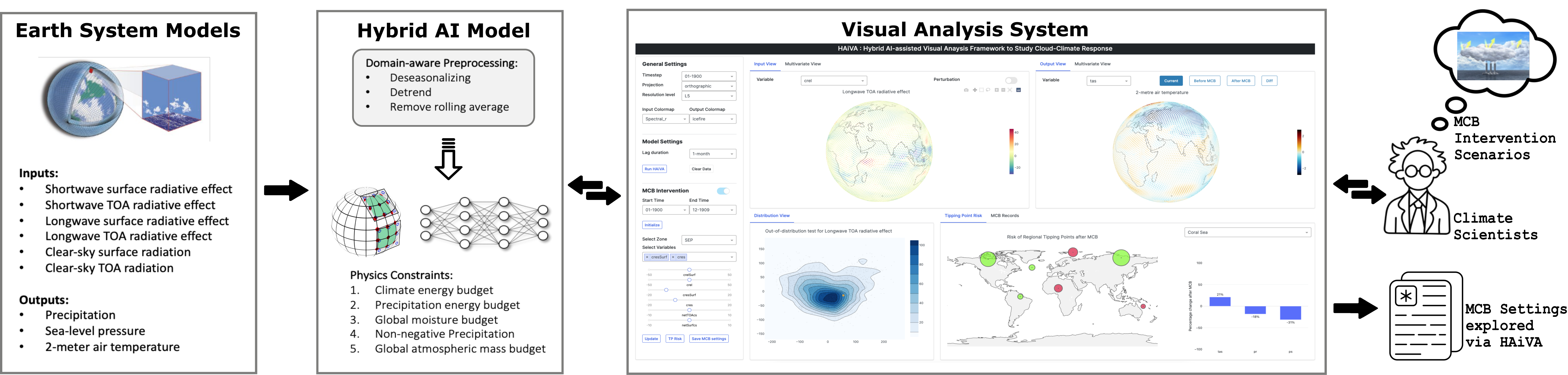}
  \caption{A high-level schematic diagram of our proposed hybrid AI-assisted visual analysis framework}
  \label{fig:teaser}
}

\abstract{Clouds have a significant impact on the Earth's climate system. They play a vital role in modulating Earth’s radiation budget and driving regional changes in temperature and precipitation. This makes clouds ideal for climate intervention techniques like \textit{Marine Cloud Brightening} (MCB) which refers to modification in cloud reflectivity, thereby cooling the surrounding region. However, to avoid unintended effects of MCB, we need a better understanding of the complex cloud to climate response function. Designing and testing such interventions scenarios with conventional Earth System Models is computationally expensive. Therefore, we propose a hybrid AI-assisted visual analysis framework to drive such scientific studies and facilitate interactive ``what-if" investigation of different MCB intervention scenarios to assess their intended and unintended impacts on climate patterns. We work with a team of climate scientists to develop a suite of hybrid AI models emulating cloud-climate response function and design a tightly coupled frontend interactive visual analysis system to perform different MCB intervention experiments.  
} % end of abstract

\begin{document}


\firstsection{Introduction}

\maketitle

%% \section{Introduction} %for journal use above \firstsection{..} instead
% This template is for papers of VGTC-sponsored conferences which are \emph{\textbf{not}} published in a special issue of TVCG.
\setlength{\belowcaptionskip}{-10pt}
Clouds play an important role in modulating Earth's climate by regulating the amount of solar energy that reaches the surface and the amount of energy that is radiated back into space. They affect the coupled circulation of atmosphere and ocean, driving regional changes in temperature and precipitation~\cite{2005cloudclimate}.  However, the climate response to changing clouds is one of the largest uncertainties in our \textit{Earth System Models} (ESM)~\cite{flato2011earth, giorgi2018regional}, particularly when producing long-term climate projections. This limitation becomes apparent when analyzing potential climate change impacts due to large changes in cloud properties, e.g., the presence of greenhouse gases (loss of clouds) or climate intervention techniques (higher cloud reflectivity)~\cite{mah2020coping, lee2019systematic, valentini2006goal}. Climate intervention techniques like \textit{Marine Cloud Brightening} (MCB)~\cite{ stjern_response_2018,latham2012marine,latham2012weakening}, which involves increasing the reflectivity of marine clouds (i.e. “brighten” them) by introducing additional sea-spray particles near the clouds, require a thorough understanding of how cloud properties impact large-scale global circulation and regional climate patterns. This is also critical to assess and avoid any unintended effects of MCB on overall climate patterns. Such studies need to be fine-tuned spatio-temporally, by running thousands of ESM ensembles to find optimal cloud brightening strategies. However, current ESM simulations are computationally expensive, requiring tens of thousands of core-hours, thus making it an impractical approach to explore a wide range of possible MCB interventions and their potential impacts. To facilitate climate scientists in this process, we have developed a hybrid AI-assisted visual analysis framework to study cloud to climate responses and interactively explore different intervention scenarios for MCB and evaluate their potential impacts across sensitive geospatial sites around the globe.

Our proposed framework comprises of a suite of hybrid AI emulator models trained for different time-lagged duration and a frontend interactive visual analysis (VA) system which uses these models in the backend to analyze the impact of cloud properties on climate patterns. The formulation of the hybrid AI modeling approach is guided by \textit{Fluctuation-Dissipation Theorem} (FDT)~\cite{kubo_fdt_1966}, a fundamental principle in statistical physics which states that the forced response of a system mirrors its internal fluctuations. Therefore, we perform physics-aware preprocessing of ESM simulation data to remove any seasonal cycles and secular trends, thus extracting month-to-month internal climate variability (i.e. climate \textit{anomalies}). The skeleton of our hybrid model is a Multilayer Perceptron (MLP) architecture, that is designed to handle geodesic data via a spherical graph sampling method. The model takes as input 6 different cloud radiative fields and predicts \textit{sea-level pressure}, \textit{precipitation}, and \textit{air temperature} as outputs. To ensure physically consistent predictions, we include soft physics constraints based on radiative forcing budget formulations during the training process.

% Our proposed framework comprises of a suite of hybrid (data-driven models with physics constraints) AI emulator models and an interactive visual analysis (VA) system which uses trained hybrid models in the backend to analyze the impact of cloud properties on climate patterns. The skeleton of our hybrid model is a Spherical U-net architecture, that is specifically designed to handle geodesic data. To ensure physically consistent model predictions, we include physics constraints (loss functions) to enforce the interplay between cloud forcing and radiative fluxes based on radiative forcing budget formulations. The models take as input shortwave and longwave cloud radiative effects as well as clear-sky radiation at both top of the atmosphere and surface level and tries to map to \textit{sea-level pressure}, \textit{precipitation}, and \textit{air temperature} as outputs.     

The frontend interactive VA system is developed following the requirements and guidelines expressed by our team of climate scientists. The multi-panel design lets the scientists explore different facets of the ESM data and corresponding model-driven analyses of cloud-climate properties. They can interactively run the hybrid AI models in the backend, visualize the input and the output anomaly fields using popular geospatial projections, as well as explore their multivariate relationships across different spatial regions using interactive parallel coordinate plots (PCP). To replicate the key steps involved in MCB intervention studies we incorporated several user-input controls in our VA system. Users can select the \textit{time-duration}, \textit{spatial region(s)}, \textit{radiation field(s) to perturb} and the \textit{extent of perturbation} for each MCB intervention scenario analyzed using our system. Before running the backend hybrid models with the new MCB input fields, scientists can validate any out-of-distribution cases by visualizing the distribution-shift of the new radiation fields. Another important aspect of this analysis is to track the potential impact of an MCB intervention on the risk of climate tipping point at key climate-sensitive geospatial site across the globe. We assess surface climate changes that indicate the tendency of MCB impacts on seven different tipping points for each MCB intervention performed in our VA system. Finally, during the analysis task, the scientists can progressively save interesting MCB intervention settings in a tabular form which can be later exported as excel/csv files for further downstream experiments and expensive simulation executions. Thus, our hybrid AI-assisted VA system serves as an efficient scientific analysis workflow for climate scientists trying to study complex cloud-climate relationships and hypothesize the impact of different MCB-based climate intervention scenarios. We validated the predicted climate responses of our hybrid model for a subset of controlled intervention scenarios against baseline ESM simulations as well as performed qualitative evaluation of the VA system with climate scientists and experts.

\begin{figure*}[t!]
\centering
        \includegraphics[width = 0.93\linewidth]{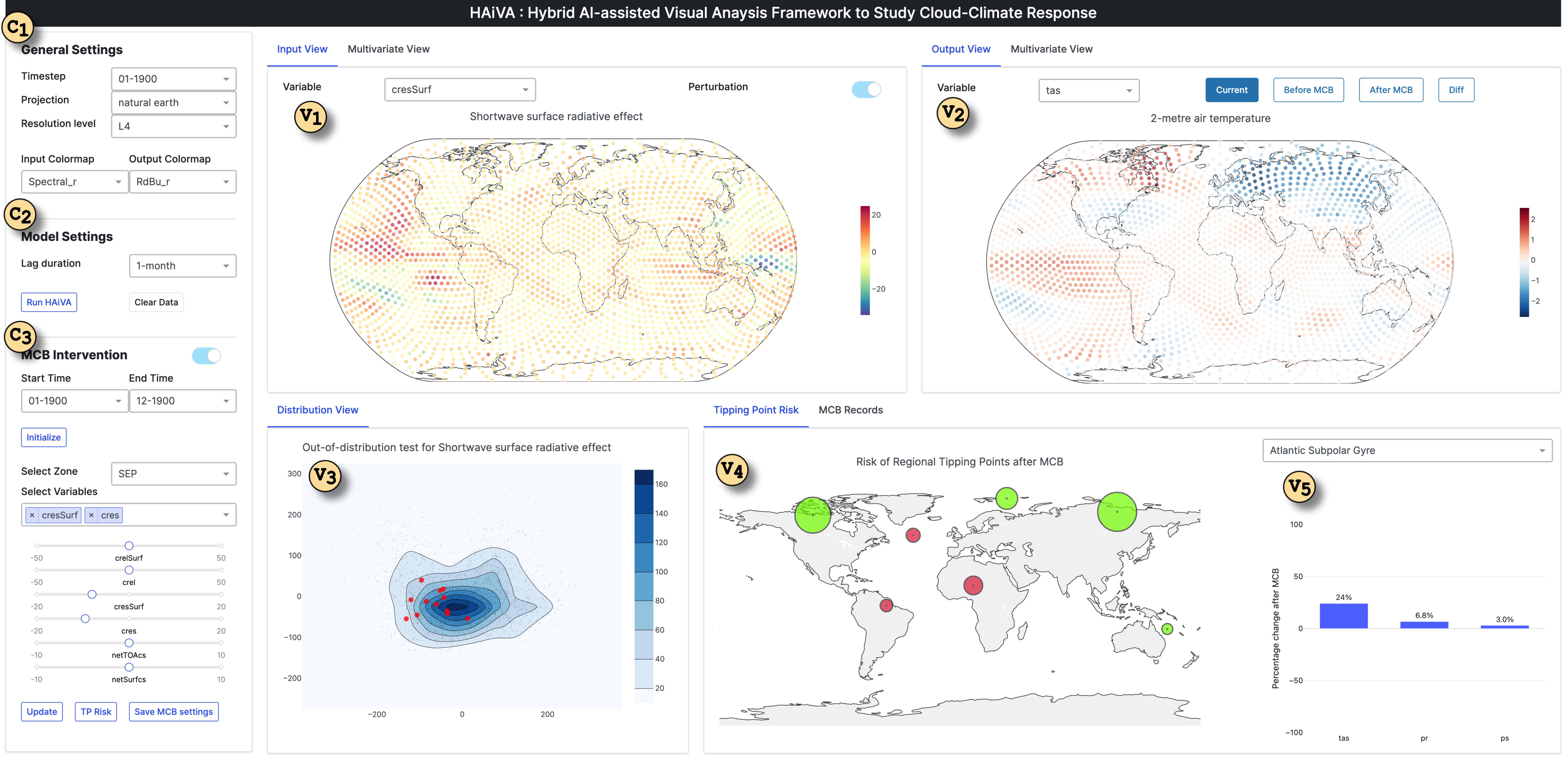} 
\caption{Complete layout of our proposed hybrid AI-assisted visual analysis system. (C$_1$, C$_2$, C$_3$) User input control panel to set analysis parameters, including MCB intervention experiments. (V$_1$) Input cloud radiation anomaly visualization panel. (V$_2$) Output climate anomaly visualization panel. (V$_3$) Distribution-shift visualization. (V$_4$, V$_5$) Tipping point risk assessment panel.}
\label{HAiVA}
\end{figure*}

\section{Related Work}
Visual analytics is an interdisciplinary field that combines the expressive power of data visualization techniques with the modeling power of statistical and machine learning methods to create efficient analysis workflows. Endert et al.~\cite{ml4va}, in their 2018 paper, provides a comprehensive survey of how this synergy between machine learning and data visualization have grown in the visualization research community.  Machine learning models like SVM (Support Vector Machine)~\cite{xie2019visual}, LDA (Latent Dirichlet Allocation)~\cite{lee2012ivisclustering,liu2012tiara}, KNN (K Nearest Neighbor)~\cite{marai2018precision}, Bayes' rule~\cite{goodall2019situ}, learning-from-crowds model~\cite{liu2019interactive}, and online metric learning~\cite{liang2018photorecomposer} have been extensively utilized by visualization researchers to enhance the analysis experience in their VA systems. Deep learning models like CNN (Convolutional Neural Network)~\cite{xie2018semantic} and Word2Vec~\cite{zhou2019visual,lu2019inkplanner} models have also been utilized to design interactive VA systems for different domain problems. Among the recent works of utilizing such system for scientific domains, graph neural network based surrogate was designed to explore the parameter space of expensive ocean simulation models~\cite{shi2022gnn}. A multilayer perceptron surrogate model was used to design interactive VA system for a computational biology domain~\cite{NNVA2019}. Our proposed framework follows similar VA system design philosophy to facilitate cloud-climate relationship analysis.

\section{Climate Science Background}
\label{background_sec}
% \subsection{Marine Cloud Brightening} 

% Climate intervention techniques refer to specific geo-engineering methods aimed at reducing the effects of climate change by manipulating Earth's environment in a measured approach. MCB along with Stratospheric Aerosol Injection (SAI) are considered as the two most feasible and practical Solar Radiation Management (SRM) strategies, which involve reflecting sunlight back into space to cool the Earth. While SAI introduces aerosol particles such as sulphates into the stratosphere, MCB on the other hand involves spraying of sea-salt aerosols into marine boundary layer clouds in the lower atmosphere to increase their albedo. Both the intervention techniques require extensive research and in-depth understanding of their intended as well as unintended impacts on the Earth system. In this work, we are collaborating with climate scientists to focus only on MCB intervention technique. The scientific community is particularly interested in understanding the impact of such interventions on different climate tipping points identified for several sensitive geo-spatial sites across the globe. \textit{Tipping points} are critical thresholds in our climate system, when crossed, can lead to large and often irreversible changes with severe impacts on human society. Collapsing of the West-Antarctic ice sheet, shift in West African monsoon, and warming of the Coral Reefs are some examples of identified tipping point risks. 

\textbf{Marine Cloud Brightening (MCB):} Climate intervention techniques essentially aim at reducing climate change impacts by manipulating Earth's environment~\cite{mah2020coping, lee2019systematic, valentini2006goal}. MCB is a notable type of Solar Radiation Management (SRM)~\cite{robock2020benefits, nicholson2018solar} strategy, a category of proposed climate intervention techniques to reflect sunlight back into space to cool the Earth. MCB involves spraying of sea-salt aerosols into lower atmosphere marine clouds to increase their albedo (cloud reflectivity), thereby cooling surrounding regions~\cite{ stjern_response_2018,latham2012marine,latham2012weakening}. Such intervention techniques require extensive research and in-depth understanding of their intended as well as unintended impacts on the Earth system. The scientific community is particularly interested in understanding the impact of such interventions on different climate tipping points identified for several sensitive geo-spatial sites around the world \cite{united_nations_environment_programme_one_2023}. \textit{Tipping points} are critical thresholds in our climate system, that when crossed can lead to large and often irreversible changes with severe impacts on human society~\cite{mckay_exceeding_2022}. For e.g.,  West-Antarctic ice sheet collapse, West African monsoon shifts, and coral reef collapses.

The effects of MCB are localized, making the possibility space of MCB intervention scenarios vast, both in terms of forcing strength, spatial patterns and temporal duration of interventions. This makes MCB an ideal candidate to perform a wide range of ``what-if" analyses to develop a thorough understanding of both the positive and negative impacts of different scenarios. However, the large number of possible scenarios also make it a computationally prohibitive task to design such an exploration tool because it would require running computationally expensive ESM simulations multiple times. ESMs are comprehensive dynamical models of the coupled atmospheric-ocean-land-ice system~\cite{flato2011earth, giorgi2018regional} and can take thousands of core-hours to run few instances, severely constraining thorough MCB impact assessment. Past evaluations of MCB have only considered a small number of controlled intervention scenarios~\cite{rasch_geoengineering_2009,jones_climate_2009,stjern2018response,kravitz2015geoengineering}. 
% To the best of our knowledge, our proposed hybrid AI-assisted visual analysis framework is the first of its kind to facilitate interactive exploration of different climate intervention scenarios and assess their potential impacts.      

\noindent
\textbf{Fluctuation-Dissipation Theorem (FDT):} A challenge towards modeling such a climate response emulator is that there are only a handful of MCB forcing ESM simulations that have been conducted, meaning we cannot train on any existing large repository of ESM responses to MCB forcing data. Thus, we formulate our modeling approach based on the fundamental principles of FDT~\cite{kubo_fdt_1966,leith1975climate}, a theorem in statistical mechanics that posits that \textit{the response of a dynamical system to a perturbation can be inferred from the time-lagged correlation statistics of natural internal fluctuations in the system}. Since climate system is hypothesized to be such a dynamical system, FDT has been used to estimate the linear response of climate to solar radiation perturbation~\cite{cionni_fluctuation_2004} and regional ocean heat convergence~\cite{liu_sensitivity_2018}. As such, we perform a domain-aware data pre-processing to extract natural internal fluctuations from ESM data and train a suite of AI models at different time-lagged duration. Applying the concepts of FDT, we time-integrate the outcome from these models to estimate the climate impact of the external radiative forcings, thus emulating an MCB intervention scenario.

% that have been conducted in the most recent generation of Coupled Model Intercomparison phase 6 (CMIP6) ESMs,

\section{Requirement Analysis}
\label{RA_section}
Throughout the course of this project, spanning over a year, we had regular weekly interactions with our team of climate scientists to understand their analysis needs and modeling requirements. These also helped us formulate the problem to best capture the physics of a cloud-climate response function. Here are some of the main requirements set-forth for our visual analysis framework:
\begin{itemize}
	\setlength\itemsep{0.10em}
	\item[\textbf{R1}]\label{R1} Ability to perform MCB intervention scenarios directly for the VA system and record/save the settings for interesting scenarios to execute downstream experiments and simulations.

	\item[\textbf{R2}] Ability to monitor any out-of-distribution instances when running the hybrid models with new perturbed input forcing fields. 
 
    \item[\textbf{R3}] Ability to assess the potential impact of an intervention on the risk of climate tipping points for selected geo-spatial sites.    
 
    \item[\textbf{R4}] Ability to execute the individual time-lagged AI models separately (outside the intervention scenarios) and analyze the cloud to climate response properties, including the various inter-variable relationships.
    
\end{itemize}

\section{Hybrid AI Model}
% Our backend hybrid AI emulator was trained to model the cloud to climate response function. 
Here we briefly elaborate on the key hybrid modeling steps and some of their underlying rationale. Additional details are provided as supplementary material. 

\textbf{Data:} Our training data consists ESM data from a state-of-the-art ESM, the Community Earth System Model 2 \cite{danabasoglu_community_2020}. As FDT requires a large amount of climate noise (i.e. internal variability) for training, we use the CESM2 Large Ensemble (CESM2-LE) data~\cite{rodgers_ubiquity_2021}. Specifically, we use the 50-member ensemble of historical simulations with smoothed biomass burning emission at nominal 1 degree spatial resolution and at a monthly temporal resolution, accounting for nearly 100,000 months of data.

\textbf{Preprocessing:} 
% To capture the internal fluctuations of the system, we preprocess the CESM2-LE data to remove the seasonal cycle and secular trends in the data. To achieve this, we compute the average across the 50 simulations (the ensemble average) for each variable, month, and spatial grid point in the data set. This ensemble average is then subtracted from each simulation to obtain the monthly fluctuations of the variables about the mean. The resulting data fields are generally referred to as \textit{climate anomalies} in the climate science community. Our model takes as input 6 different \textit{radiation anomalies} which include both shortwave and longwave cloud radiative effects at both top of the atmosphere (TOA) as well as surface-level along with clear-sky radiation fields. The model predicts 3 different \textit{climate anomalies} (sea-level pressure, precipitation, and temperature) as output. 
To capture the internal fluctuations of the system, we perform specific physics-aware preprocessing steps. We \textbf{\textit{de-seasonalize}} the data to remove any trends due to annual seasonality by substracting climatologoical seasonal cycles from monthy data. We \textbf{\textit{de-trend}} by fitting a third degree polynomial for each month of the year and subtract it from the data to remove secular trend in data over time. Finally we \textbf{\textit{remove rolling average}} by calculating anomaly at each grid point relative to a running mean that is computed over a 30-year window for that grid point and month. The resulting data fields are generally referred to as \textit{climate anomalies} in the climate science community. Our model takes as input 6 different \textit{radiation anomalies} which include both shortwave and longwave cloud radiative effects at both top of the atmosphere (TOA) as well as surface-level along with clear-sky radiation fields. The model predicts 3 different \textit{climate anomalies} (sea-level pressure, precipitation, and temperature) as output.

\textbf{Spherical Grid Sampling:} ESM data is typically stored on a regular 2-D latitude-longitude grid format, which have non-uniform area coverage over the globe. To incorporate the inherent rotational symmetry of the Earth we utilize a geodesy-aware spherical sampling approach~\cite{defferrard2020deepsphere} to convert the 2-D regular grid to a spherical 1-D icosahedral mesh. The vertices of the icosahedron are equally spaced points over the sphere that circumscribes it. This sampling technique also naturally provides a hierarchical multi-resolution structure, which lets us filter the spatial data at different resolution levels. We utilize this feature to offer a scalable and interactive visual analysis experience for the end-user. We trained our models at the highest resolution level (L5: 10242 grid points with $\sim$100km physical resolution) but during interactive visual analysis phase, depending on the computational resource of the analysis machine, users can select any desired low-resolution levels for exploration.

% \textbf{Spherical Sampling:} ESM data is typically stored on a regular latitude-longitude grid format, which have non-uniform area over the globe with smaller areas at the poles and larger areas at the equator making it challenging to assign equal importance during model training as well as incorporate the inherent rotational symmetry of Earth. Therefore, we utilize a geodesy-aware spherical sampling approach~\cite{} to convert the 2-D regular grid to a spherical 1-D icosahedral mesh. The vertices of the icosahedron are equally spaced points over the sphere that circumscribes it. This sampling technique also naturally provides a hierarchical multi-resolution structure, which lets us filter the spatial data at different resolution levels. We utilize this feature to offer a scalable and interactive visual analysis experience for the end-user. We trained our models at the highest resolution level (L5 with 10242 grid points) but during interactive visual analysis phase, depending on the computational resource of the analysis machine, users can select any desired low-resolution levels for exploration. 

\textbf{Multilayer Perceptron (MLP):} The architecture of our main machine learning model is an MLP network which takes in the full icosahedral (L5) grid points and considers the input and output anomalies as respective channels. Our MLP architecture has four hidden layers, each containing 1024 neurons. Layer normalization~\cite{ba2016layer} and Gaussian Error Linear Units (Gelu) activation~\cite{hendrycks2016gaussian} were employed in each layer. The learning rate was initially set to 2x$10^{-4}$ and exponentially decayed at a rate of 1x$10^{-6}$ per epoch. The model have 108M trainable parameters, and was trained for 15 epochs on the historical CESM2-LE dataset. For training we used a single NVIDIA Tesla V100-SXM2 GPU using 16GB VRAM. We trained multiple such models for different monthly time-lagged duration. As explained in Section~\ref{background_sec}, following the modeling principles of FDT, we can use these time-lagged model outputs to generate a time-aggregrated result which emulates the forcing effect of an MCB intervention for long time period.

\textbf{Physical Constraints:} To accelerate training and prevent the model from learning physically irrelevant statistical relationships, we impose a set of four soft physics-informed constraints: \textit{non-negative precipitation}, \textit{atmospheric moisture budget}, \textit{atmospheric mass budget}, \textit{atmospheric energy budget}. These constraints are implemented as additional weighted terms along with the original mean-squared error (MSE) loss function, thus, penalizing large violations of natural conservation laws during model training. 

% Further details about the models and different design and experiment details are provided in the accompanying supplementary material.

\section{Visual Analysis System}
% \begin{figure}[t!]
% \centering
%         \includegraphics[width = 0.98\linewidth]{figures/pcp_interaction.pdf} 
% \caption{Interactive PCP based visualization in the ``Multivariate View" of our system to explore the inter-variable relationships. Users can select a specific spatial region of interest by filtering out the latitude and longitude ranges to view a local multivariate properties. }
% \label{pcp_view}
% \end{figure}

% \begin{figure}[t!]
% \centering
%         \includegraphics[width = 0.80\linewidth]{figures/freeform_selection.png} 
% \caption{Freeform selection of arbitrary spatial sites to perform MCB intervention experiments using our system.}
% \label{freeform}
% \end{figure}

\begin{figure*}[t!]
\centering
        \includegraphics[width = 0.97\linewidth]{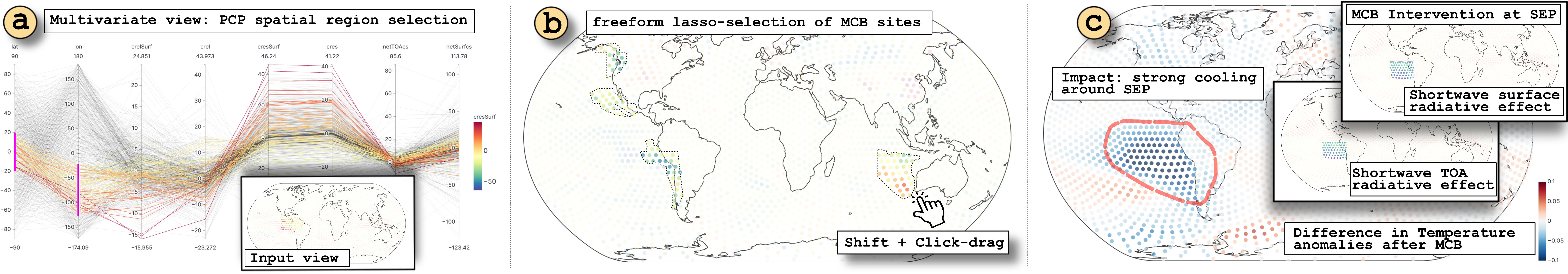} 
\caption{(a) Example of multivariate relationship exploration for different spatial regions using interactive PCP to filter lat-lon ranges. (b) Freeform selection of arbitrary spatial sites to perform MCB intervention experiments using our system. (c) A concise view of the results of MCB intervention over SEP region for 10 year duration leading to strong cooling effect in the Pacific.}
\label{fig2}
\end{figure*}

Our frontend VA system adopts a multi-panel inter-linked layout design as shown in Figure~\ref{HAiVA}. Here we elaborate on the different interactions and visualization features incorporated in our system to meet the analysis requirements set-forth by the climate scientists.

\textbf{Interaction Panel:} All the major user interaction settings and parameters needed to conduct detailed analysis are put together in an interaction panel on the left of the system. Figure~\ref{HAiVA}(C$_1$, C$_2$, C$_3$) highlight the general analysis settings, backend hybrid model settings, and MCB intervention experiment settings respectively. \textit{``General Settings"} in C$_1$ lets the users select the desired timestep from the loaded dataset, specific projection scheme for visualizations (we include 22 popular geospatial scheme options), the resolution level to use for visualization (5 levels of resolution), and the colormaps to use for the input and output variable visualizations (i.e, in V$_1$ and V$_2$ respectively). \textit{``Model Settings"} in C$_2$ lets the user select which time-lagged model to use in the backend, directly run the AI emulator model and also clear any previous prediction data from memory. C$_1$ and C$_2$, therefore addresses most of the \textbf{R$_4$} requirements. To start performing MCB intervention experiments and conduct different ``what-if" analyses users have to first activate the C$_3$ panel. This lets us select the time duration for intervention, which spatial site to perform intervention over, which input cloud radiation anomaly fields to perturb (can select multiple fields), and corresponding sliders to set the extent of perturbation. For MCB site selection, currently we offer the option to select from 3 specific sites (Southeast Pacific, Southeast Atlantic, and Northeast Pacific) that the scientists are interested in and have conducted past studies over. There is also a \textit{``freeform"} spatial site selection option which uses interactive lasso-based tools to select any arbitrary spatial region. Therefore, C$_3$  supports most of the requirements set-forth in requirement \textbf{R$_1$}. 

% Finally users can apply these MCB settings by running the backend hybrid AI models with the new perturbed input cloud radiation fields.

\textbf{Cloud Radiation and Climate Field Visualization:} The 6 input cloud radiation anomaly fields and the 3 output climate anomaly fields are visualized in the \textit{``Input view"} (Fig~\ref{HAiVA} V$_1$) and \textit{``Output view"} (Fig~\ref{HAiVA} V$_2$) panels respectively. Users can select the specific variables to visualize from the dropdown box in both the panels. V$_1$ has an additional toggle switch to view the corresponding perturbed radiation anomaly fields, useful during MCB intervention experiments. V$_2$ has four additional options to select which set of output fields to visualize. \textit{``Current"} refers to the model output for the currently selected timestep (set in C$_1$), \textit{``Before MCB"} and \textit{``After MCB"} show the time aggregated outputs without intervention and with intervention respectively. \textit{``Diff"} refers to the difference in the outputs from \textit{``Before MCB"} and \textit{``After MCB"}, which is useful to track the overall impact of the intervention experiment. Both V$_1$ and V$_2$ has additional \textit{``Multivariate View"} tabs to help study the multivariate relationships among the respective anomaly fields. Parallel coordinate plots (PCP) are provided to interactively explore the multivariate relationships across the spatial domain. The latitude and longitude information is also plotted along two separate axes to let the users interactively filter out different spatial regions, as illustrated in Figure~\ref{fig2}a. Moreover, there are additional lasso-based interaction features incorporated in V$_1$ to select arbitrary MCB sites for intervention as shown in Figure~\ref{fig2}b. To activate this feature users have to select the \textit{``freeform"} option in \textit{``Select Zone"} dropdown in panel C$_3$. Therefore, most of the requirements associated with both \textbf{R$_1$}  and \textbf{R$_4$} are addressed by the V$_1$ and V$_2$ panels. 

\textbf{Distribution-shift Visualization:} Performing MCB intervention experiments involve providing a new perturbed input field to the hybrid AI models to make new predictions. Despite the fact that we have incorporated the physics behind cloud-climate response function during overall emulator modeling, there can be a possibility that the model may not behave well when extreme out-of-distribution radiation fields are provided as input. To track such unusual extrapolated MCB scenarios we provide distribution-shift visualization as highlighted in Figure~\ref{HAiVA}(V$_3$). For every MCB intervention experiment, we plot the perturbed input anomaly fields (marked in red dots) against the distribution of the original radiation fields (blue density plot) from the training dataset. To generate this distribution-shift visualization, we project the spatial fields to lower dimensions using Principle Component Analysis (PCA). The density plots are created using kernel density estimate (KDE) on the PCA space, thus capture the high-dimensional distribution shape. V$_3$, therefore directly addresses the requirement in \textbf{R$_2$}.     

\textbf{Tipping Point Risk View:} Given the complex climate teleconnection within the Earth system, it is important that for any climate intervention scenario we assess their potential impact on sensitive geospatial sites across the globe~\cite{mckay_exceeding_2022}. We look at climate anomaly metrics that are indicators of potential changes to climate tipping point risk for 7 sites for every MCB experiment performed in our system via the panels V$_4$ and V$_5$ as shown in Figure~\ref{HAiVA}. V$_4$ shows the world map marking those 7 sites. The size of the dots represent the average spatial radius of the region. Each site has different metrics for tipping point risk assessment based on the 3 climate variables that our model predicts. If because of the MCB experiment, some of these conditions are met, we mark that site in red indicating a risk of reaching climate tipping point because of the performed MCB intervention, else color green is used to indicate no such risk for the site. We display the full condition for each site as a pop-up on mouse-hover. The detail percentage increase/decrease of the 3 climate anomaly fields for the sites are shown on the right as bar-charts in V$_5$. This panel addresses the specific requirement in \textbf{R$_3$.}   

\textbf{MCB Intervention Records:} The goal of this framework is to create a scientific analysis workflow to study cloud-climate response and hypothesize different MCB intervention scenarios (\textbf{R$_1$}). As such, we allow the users to progressively store interesting MCB intervention settings like MCB site location, perturbation parameters and duration in a tabular form (\textit{``MCB Record"} tab in Figure~\ref{HAiVA} V$_4$). Users can include additional comments/observations as notes for each entry, which can be later be exported as excel/csv files for further downstream experiments and expensive simulation execution.

\section{Discussion}
\textbf{Validation and Case Studies:} To validate that we can plausibly project the climate response to MCB-like perturbations, we compared responses from a novel set of fully coupled CESM2 simulations~\cite{hirasawa_impact_nodate} to those of time-integrated responses of our hybrid model for radiative flux anomalies computed from fixed sea-surface temperature MCB simulations. Three previously studied spatial regions were included in these validation case studies: Northeast Pacific (NEP), Southeast Pacific (SEP), and Southeast Atlantic (SEA). Scientists found that our hybrid AI emulator was able to reproduce the patterns of climate response to MCB with strong correlation scores for all the three anomalies. They were also able to successfully observe key remote teleconnected responses to the MCB forcings, specifically La Ni\~na-like temperature signals, with strong cooling in the tropical Pacific and warming in the midlatitudes east of Asia and Australia, as well as cooling over low-latitude land regions. Figure~\ref{fig2}c shows a concise view of one such study over the SEP regions for a 10 year duration using the VA interface, showing a strong cooling effect around the surrounding region. We provide a supplementary video to demonstrate such MCB intervention cases in detail and highlight all the interaction features. The system also highlighted key precipitation changes like drying in northeast Brazil, central Africa, and southern North America, and wetting in the Sahel, southeast Asia, Australia and central America. Our scientists plan to report such interesting observations, indicating the ability of hybrid AI models to correctly project climate responses to MCB forcing in a climate science journal in near future.

\noindent
\textbf{Expert Feedback:} Our team of scientists comprises of two climate science professors and two postdoctoral researchers. They feel that the final VA system meets all the requirements to conduct interactive MCB-type climate intervention scenario testing. They found the user interface to be intuitive enough to replicate intervention studies. Outside this group of experts, we presented our work to several climate stakeholders and scientific program officers. They strongly feel that such an interactive VA system can not only assist core climate science studies of future intervention techniques but can also serve as a tool for public and government engagement in several climate change initiatives. Concrete examples were drawn on how it can be used by policy makers to make climate impact-aware geo-political decisions. 

\noindent
\textbf{Implementation:} The VA system was developed using Plotly's Dash platform, which utilizes Flask as the backend engine to run the trained hybrid models, React.js for user-input controls and Plotly.js for rendering visualization. Additional details and in-depth explanation for different stages of our work can be found in our parent project's open-source documentation page~\cite{aibedo}. 

\section{Conclusion and Future Work}

In this paper, we have proposed a hybrid AI-assisted visual analysis framework to study the complex cloud-climate response function and facilitate rapid exploration of a wide range of MCB intervention scenarios along with their potential climate change impacts. To the best of our knowledge, this is the first interactive VA system to study any form of climate intervention technique, an active area of climate science research. In future, we would like to update our backend models to get better predictions for high-latitude regions. This will help evaluate key crysopheric tipping points such as Eurasian and North American permafrost loss. We would also like to incorporate predictive uncertainty quantification to our hybrid model outcomes.

\acknowledgments{
This works is part of the AIBEDO project, funded under the DARPA AI-assisted Climate Tipping-point Modeling (ACTM) program under award DARPA-PA-21-04-02.}

\bibliographystyle{abbrv}

\bibliography{template}
\end{document}